\documentclass[acmsmall]{acmart}

\pdfoutput=1

\AtBeginDocument{%
	\providecommand\BibTeX{{%
			\normalfont B\kern-0.5em{\scshape i\kern-0.25em b}\kern-0.8em\TeX}}}

\renewcommand\footnotetextcopyrightpermission[1]{} 
\newcommand{\bem}[1]{\textbf{\emph{#1}}}

\begin{document}
	\title{Privacy Guidelines for Contact Tracing Applications}
	
	\author{Manish Shukla}
	\email{mani.shukla@tcs.com}
	\affiliation{%
		\institution{TCS Research}
		\city{Pune}
		\country{India}}
	
	\author{Rajan M A}
	\email{rajan.ma@tcs.com}
	\affiliation{%
		\institution{TCS Research}
		\city{Bangalore}
		\country{India}}
	
	\author{Sachin Lodha}
	\email{sachin.lodha@tcs.com}
	\affiliation{%
		\institution{TCS Research}
		\city{Pune}
		\country{India}}
	
	\author{Gautam Shroff}
	\email{gautam.shroff@tcs.com}
	\affiliation{%
		\institution{TCS Research}
		\city{Delhi}
		\country{India}}
	
	\author{Ramesh Raskar}
	\email{raskar@mit.edu}
	\affiliation{%
		\institution{MIT Media Lab}
		\city{Cambridge, MA}
		\country{USA}}

\begin{abstract}
	\bem{Abstract.} \emph{Contact tracing is a very powerful method to implement and enforce social distancing to avoid spreading of infectious diseases. The traditional approach of contact tracing is time consuming, manpower intensive, dangerous and prone to error due to fatigue or lack of skill. Due to this there is an emergence of mobile based applications for contact tracing. These applications primarily utilize a combination of GPS based absolute location and Bluetooth based relative location remitted from user's smartphone to infer various insights. These applications have eased the task of contact tracing; however, they also have severe implication on user's privacy, for example, mass surveillance, personal information leakage and additionally revealing the behavioral patterns of the user. This impact on user's privacy leads to trust deficit in these applications, and hence defeats their purpose.} 
	
	\emph{In this work we discuss the various scenarios which a contact tracing application should be able to handle. We highlight the privacy handling of some of the prominent contact tracing applications. Additionally, we describe the various threat actors who can disrupt its working, or misuse end user's data, or hamper its mass adoption. Finally, we present privacy guidelines for contact tracing applications from different stakeholder's perspective. To best of our knowledge, this is the first generic work which provides privacy guidelines for contact tracing applications.}
	
\end{abstract}

\maketitle

\pagestyle{empty}

\section{Introduction}
	Last two decades have witnessed multiple occurrences of infectious diseases such as SARS, MERS, HIN1, Ebola and SIV. The current outbreak of \emph{COVID-19} has been declared as a pandemic and public health emergency of international concern by World Health Organization (WHO) \cite{who2020pandemic}. Outbreak of such diseases have severe effect on a nation's economy and health infrastructure. Additionally, it induces fear in citizens and creates bias towards individuals with infection that further leads to their discrimination \cite{person2004fear}. This discrimination further impacts the businesses associated with the infected individuals \cite{raskar2020apps}.
	
	The \emph{COVID-19} and \emph{influenza} viruses are similar in some ways as both cause the respiratory disease with no symptom or mild to severe conditions and in severe cases they may result in death. Further, their mode of transmission consists of contact, droplets and fomites. However, the basic reproduction number ($R_0$) of \emph{COVID-19} is greater than that of \emph{influenza} \cite{who2020covidinf}. Here, $R_0$ is the average number of people who will catch the disease from the infected person, hence a higher value signifies that \emph{COVID-19} is more contagious than \emph{influenza}. Additionally, the mortality rate of \emph{COVID-19} appears to be higher than that of \emph{seasonal influenza}, although, the rate also depends on access to quality health services. 
	
	Kelso et al \cite{kelso2009simulation} has shown that social distancing is an effective measure against a virulent strain of an infectious disease with no vaccine. They further recommended that for a given disease, the social distancing should be activated without delay (considering its $R_0$ value) and for a relatively long period. When the virus has high mortality rate then the social distancing is implemented as a complete lockdown, which requires closing roads, restricting travel, and community-level quarantine. In case the transmission has age or gender bias or if it is a more infectious strain, then more restrictive steps need to be taken, for example, offices and workplaces must also be closed down \cite{glass2006targeted}. The lockdown approach is useful in limiting the spread of infectious diseases; however it also has a long-term detrimental effect on a nation's economy \cite{jones2020ecoimpact,imf2020lessons}. Also, the rising unemployment and social isolation due to lockdown gravely impacts an individual's health in various ways \cite{pell2020healthimpact,raskar2020apps}. Furthermore, it is very difficult to answer `what should be the correct duration for lockdown?' as it depends on multiple factors, for example, demography, availability and quality of health services, population density, economic condition and political will power. This implies that we require an alternate method which eases the pressure on available infrastructure and humans alike. 	
		
	Contact tracing is a well-studied method for curbing the spread of infectious diseases. It requires modeling the potential transmission routes through which an individual could infect the others \cite{eames2003contact}. This enables authorities in isolating the suspected and confirmed cases for further observation or treatment. This eases the load on already overstretched infrastructure. Moreover, it also makes individuals aware about the situation and pushes a potentially infected person to self-quarantine \cite{raskar2020apps}. This in turn effectively helps in reducing the $R_0$ value. Traditionally contact tracing involves manual follow up of the known patients and then interviewing them for identifying the places that they visited and the individuals with whom they had contact in last few days. These identified individuals with asymptotic or mild symptoms are then isolated for further observation and containment. The process is cumbersome, error prone and potentially dangerous for the person performing the activity \cite{raskar2020apps}. This however has changed in the current \emph{COVID-19} outbreak, wherein different nations have adopted varied ways of contact tracing using mobile devices or wearables or both. 
	
	At some point nations are expected to ease in social distancing as a prolonged lockdown has adverse effects on economy and society. Further there is no clarity about availability of a vaccine for \emph{COVID-19} for citizens as development, trials and approvals from concerned authorities usually take time. This relaxation in lockdown can be facilitated by an effective contact tracing solution. To this end, we consider the following three scenarios for a contact tracing application, which is an extension of \cite{raskar2020apps}:
	\begin{enumerate}
		\item \bem{Lockdown Period.} It describes the duration during which social distancing is enforced by the government to curtail the spread of infectious disease by: 
		\begin{itemize}
			\item enrolling individuals 
			\item identifying contacts of infected individuals and informing them, and
			\item monitoring and follow-up
		\end{itemize}
		\item \bem{Staggered Movement.} During this period a nation could allow a part of the population to operate normally for providing essential services. All the required safeguards would have to be taken as individuals would be interacting and working together. The tracker application should help the population by providing sufficient insights and nudges to avoid any accidental outbreak. 
		\item \bem{Post Lockdown.} In case of a highly infectious disease there is always a chance of recurrence of the disease when the aggressive countermeasures are removed. Especially in case when there is an absence of herd immunity against the disease \cite{leung2020firstwave}. The tracing application should be able to provide useful insights for limiting disease recurrence. 
	\end{enumerate}
	
	For providing useful insights, contact tracing applications primarily utilize a combination of the following three data-sources \cite{openminded2020privacy}:
	\begin{itemize}
		\item \bem{Absolute location} of the individual based on device GPS data, cell phone towers, Wi-Fi routers, historical location data from a third-party service provider, like Google.
		\item \bem{Relative location} with respect to other individuals in proximity by utilizing device Bluetooth data.
		\item \bem{Group identity} based on the combination of above two and data from the additional sensors on the device. 
	\end{itemize}
	These applications are helpful in curbing the spread of disease, however they also suffer from severe privacy issues, for example, these applications could expose personal information, reveal behavioral patterns, and can also be used for mass surveillance \cite{raskar2020apps}. Researchers have shown that data privacy plays a crucial role in adoption of location-aware pervasive services \cite{chang2007user}. A survey quoted in \cite{pentina2016exploring} suggests that for 85\% of adults it is important to have control over personal their data, 54\% of users refused to download the application and 30\% of them uninstalled the application because of privacy reasons. This sensitivity towards data privacy is more profound if the application also deals with user's health data \cite{adhikari2014security}. Recently, due to these reasons, a \emph{COVID-19} tracking application from Iran has been removed from the Google Play Store for allegedly spying on its citizens \cite{cimpanu2020iran}. 
	
	There are some recent initiatives to address the privacy concern by either processing the data locally on the device \cite{gds2020tracetogether}, or using techniques like private set intersection \cite{berke2020assessing,demirag2020tracking}, or recommending usage of differential privacy \cite{cho2020contact}. In this work we describe the different scenarios which a contact tracing application should consider. Further, we discuss the various threat actors that might corrupt the data or leak sensitive data. Finally, we expand and categorize the privacy concerns of different stakeholders of \cite{raskar2020apps}. Our proposed guidelines are not specific to any particular application, but rather applies to all contact tracing application.
	
	\section{The Current Landscape}
	The existing contact tracing solutions utilize a combination of GPS and Bluetooth for determining an individual's absolute and relative position with others. Some of these solutions claim to be privacy preserving, whereas others are alleged to be coercive, poorly implemented or being used for electronic surveillance. Mobile phones have been proposed as a tool to implement a digital form of contact tracing that does not rely on memory of past social interactions. This works by having phones collect and report data that encodes when all pairs of people were in close proximity, and using that data to assess individuals' risk of contagion \cite{reardon2020surveillance}. In this section we will discuss about some such prominent applications and also the related literature.
	
	\subsection{Contact Tracing Applications} \label{sec:cta}
	\bem{China.} The contact tracing is primarily provided by Alipay and WeChat mobile applications. These applications use self-reported data by the user, their travel history, health status and government records for assigning green, yellow and red colors. The color signifies whether the user is healthy (green) or suspected of infection (yellow) or is a confirmed patient of \emph{COVID-19} (red) \cite{davidson2020chinatracing}. The system is highly coercive in nature as the color codes decide the freedom of movement of users, and also violators are threatened with severe penalties. From privacy perspective it is very intrusive as these applications require user's identity, address and travel history. Further, the color assignments are claimed to be ambiguous and there are evidences that these applications feed data back to the authorities, which could be used for surveillance \cite{knight2020chinatracing}. 
	
	\bem{Iran.} The contact tracing application from Iran required a user to register using their mobile number. It uses unencrypted GPS data for identifying the user's absolute location for contact tracing. In addition, the application also requests for an unrelated permission for identifying a user's activity. Security experts have discovered that it also sends the self-declared attributes of the user, for example, gender, name, height, and weight of a user, to the developer's server \cite{chrysaidos2020irantracing}. The application has been removed from the Google Play Store, although it is still available from other application stores \cite{cimpanu2020iran}.
	
	\bem{Singapore.} The official contact tracing application, called TraceTogether, utilizes Bluetooth data for determining the others in proximity of the user. Location permission is used only for finding the relative distance between the users \cite{gds2020tracetogether}. It requires the user to register with her phone number for fast communication. Each device periodically generates a random unique identifier for communication with nearby devices. In case a user gets infected with \emph{COVID-19} then they upload their logged data with government, which in turn broadcast it to other users for matching. TraceTogether is relatively user privacy friendly as no identifiable information is shared between the devices and the government only knows about the phone number of the registered users \cite{gds2020tracetogether,berke2020assessing}.
		
	\bem{Israel.} The `Hamagen' contact tracing application collects location history of the user in the background. It compares the user's movement with the health ministry's data on locations of confirmed \emph{COVID-19} cases. If a user has come in close contact of an infected person then an alert message is shown to the user and they are directed to a website containing further precautions to take. As per application developers, no information is uploaded on the server and all the information is stored on the phone; even the path comparison happens locally \cite{winer2020israeltracing}. This is in contrast to the permission given to `Shin Bet' security agency which allows movement tracking of smartphone users through their devices for limiting \emph{COVID-19} spread.
	
	\bem{India.} Government of India (GoI) has recently unveiled its contact tracing application called `Aarogya Setu'. The application requires only phone number for registration, although it also captures additional information, for example, name, age, sex, profession and countries visited in the last 30 days. A unique identification number, generated using phone number, is used for tracing all the contacts in close proximity of a \emph{COVID-19} patient in last 14 days. The government claims that all the captured data is stored on the local device and will be uploaded to government server for further processing. It is however not clear whether the analytics will happen on plain data and the results will be anonymized, or the analytics will directly happen on anonymized dataset. Further, a user can delete her account but the data will remain with GoI for 30 days before it will be purged. Also, any insights generated using the user's data will remain with the government \cite{agrawal2020indiatracing}. 
	
	\bem{USA.} There are multiple contact tracing applications that have emerged recently. `Covid Watch' and `CoEpi' applications use Bluetooth for proximity based contact tracing in a decentralized setup. Both these applications measure the signal strength for estimating the contact distance with other users. If the user remains in contact for a predetermined amount of time then all the devices in proximity will generate a unique `contact event number' for sharing, which is time limited and stored on the local device. When a user is diagnosed to be infected then the list of unique `contact event number' in her personal device is publicly made available with her consent. Other devices then use this public list to find out whether they came in close contact of an infected person or not. These application also allow users to locally store the location information for their own reference. This is good from an individual's privacy perspective, however will fail to inform a healthy person who never came in close contact of the infected, but rather visited the same location little later. 
	
	On the other hand, MIT's `SafePaths' is technology agnostic and utilizes multiple methods for contact tracing, for example, it uses Bluetooth, WiFi SSID and GPS data for absolute and relative location tracking of users. In case a user is found infected then her location trail is redacted by a health officer for any personally identifiable information. The redacted trail is then used for alerting other users who came in contact with the infected person \cite{berke2020assessing}. Dependency of `SafePaths' on a human to redact sensitive data could create problem if the human-in-the-loop is incompetent or has malicious intent.
	
	\bem{Others.} Other nations are also developing their own version of the contact tracing application. Majority of them are utilizing a combination of self-reporting, GPS based location tracking and Bluetooth data. To address the privacy concerns they rely on self-reporting, processing anonymized or encrypted data and preferring decentralized computation in comparison to centralized computation \cite{gdprhub2020resttracing}.

	\subsection{Related Work}  
	As contact tracing requires an individual to share their device's absolute and relative locations, therefore, it makes user susceptible to various privacy attacks, for example, real-time location surveillance and group membership discovery. Some of the contact tracing applications, mentioned in the previous section, already have some level of privacy preserving analytics \cite{gds2020tracetogether}.
	
	Raskar et al \cite{raskar2020apps} have discussed the privacy concern of stakeholders, utility versus privacy trade-off and challenges to implement such a solution. They further discussed their solution `Private Kit: Safe Paths' and the privacy measure that they have taken. Demirag and Ayday \cite{demirag2020tracking} described a private set intersection based technique for allowing user to get notified in case they have come in close contact of a diagnosed patient. They further evaluated the possibility of malicious users that may try to learn the diagnosis of some known users by tampering their local contact history. In their work, Berke et al \cite{berke2020assessing} have proposed a private set intersection protocol for assessing the risk exposure of a user to an infectious disease using anonymized and encrypted GPS locations. Oliver et al \cite{oliver2020mobile} discuss the importance of mobile data for combating \emph{COVID-19} pandemic and its effectiveness in implementing social distancing. Authors further identify the key gaps in adoption of mobile based data gathering and contact tracing, and ways to overcome these gaps. Researchers in \cite{cho2020contact}, discuss the privacy implications of the existing contact tracing applications and suggest ways to mitigate these concerns without affecting the usefulness of such applications. Similarly, a European coalition of researchers are working on a contact tracing application which complies with their strict privacy rules \cite{lomas2020europrivacy,gdprhub2020resttracing}.
	
	\section{Privacy Threat Actors}
	For designing a privacy aware contact tracing application, it is important to understand the different threat actors that may disrupt its working or impede its large scale adoption. Following is a list of privacy threat actors:
	\begin{itemize}
		\item \bem{Application Developer.} An application developer could perform the following malicious activities:
		\begin{itemize}
			\item Accessing or uploading raw device data or data of other users in proximity.
			\item Snooping on data from other applications on the device.
			\item Requesting permissions which may result in surveillance of user or her contacts.
			\item Analyzing data for generating insights which were not part of the privacy or service statement of the application.
			\item Selling data to third party.
		\end{itemize} 
	 
		\item \bem{Government.} The government could perform the following malicious activities:
		\begin{itemize}
			\item Performing selective analysis of a user, group or community.
			\item Retaining personal and discriminative data of a user even after the application is deleted by the user or the data retention period is over.
			\item Mass surveillance.
			\item Analyzing data for generating insights which were not part of the privacy or service statement of the service.
		\end{itemize} 
		\item \bem{Internet Service Provider.} The ISPs could perform all the malicious activities of `Application Developer' and `Government' if the communication between the application and the server is not protected by encryption and anonymization.		
		\item \bem{Other Applications on the User Device.} Based on their privileges, other applications can either read the locally stored data or monitor network activity or both. Their capability is similar to that of `Internet Service Provider' if the data is not properly encrypted and anonymized.
		\item \bem{Paired Devices.} These are the devices which come in the Bluetooth proximity of the user's device. If the communication between the devices is not protected by encryption then they can discover sensitive information about the nearby users. 
		\item \bem{Snooping Devices.} Same as the `Contacts' but without the contact tracing application on their device.
		\item \bem{Hackers.} Any person who accesses the user's device without their consent and proper authorization. Their capabilities are similar to that of the `Other Applications'.
	\end{itemize} 
	
	\section{Privacy Guidelines}
	There are multiple stakeholders who gets affected because of privacy violation by a contact tracing application. The impact on these stakeholders may vary from facing harsh social stigma to loss of business to restricted freedom of movement \cite{raskar2020apps,davidson2020chinatracing}. Additionally, the government (application developer) has to provide a well-defined process for taking an informed consent from a non-user for uploading their geolocation. Their location data can be taken from other applications or services installed on her device for making the risk scoring more reliable. To address this, we propose a set of guidelines that consider the privacy of the user, privacy of the infected person, privacy of the non-user and risk to the reputation of the businesses involved.

	\subsection{Guidelines for the User Privacy}
		\subsubsection{Related to Personal Data}
			\begin{itemize}
				\item The application should not ask for any personally identifiable information for downloading and availing the contact tracing services. 
				\item The application should provide a local login interface for preventing unauthorized access to data and application. 
				\item Users should never have to reveal anything about themselves unless the user is a confirmed patient.
				\item No personal data should be uploaded to the server, however if a user is found to be infected then authorities should upload the anonymized data with user's consent and after explaining the purpose for which it will be used.
				\item Any kind of device data (GPS, Bluetooth etc.), in raw or processed form, should not be sent to the server.
				\item No identifiable data from one device should go to another device, for example, Bluetooth packet containing PII.
				\item Only the GPS trail of an infected individual is uploaded to the central database to augment scoring for other users on their devices. A normal user's data should not go to the server because somehow it becomes part of someone else's interaction graph.
				\item The application should provide sufficient and necessary safeguards for protecting data stored locally or in transit or stored on server.
				
				\item The application should explicitly disclose and explain the purpose of permissions requested for its working. 
				\item The application should anonymize the data before analyzing.
				\item The application should pad or ignore the data if the data is too sparse so as to uniquely identify the user or their affiliation.  
			\end{itemize} 
		\subsubsection{Informed Consent}
			\begin{itemize}
				\item If the user has to reveal her personal data, then it should be with her informed consent and it should not be coercive (refuse medical care or punishment). However, incentives are fine (less time spent on personal interviews, faster results, insurance incentives, etc.).
				\item What data is being collected has to be made transparent to the user, and clear, explicit and affirmative consent is to be collected for that.
				\item There might be legal provisions for government agencies to lawfully collect PII information without user consent. In that scenario, the application should clearly notify the end user about it, purpose for collection and retention period etc.
			\end{itemize} 
	
	\subsection{Guidelines for the Patient Privacy}
		\begin{itemize}
			\item Minimally identifiable data from an infected person (`what') is uploaded. Such an upload should come from a health professional (`who') after the test outcome is positive (`when').
			\item Symptom reporting should not go to the server in the raw format.
			\item Infected individual's identity and location should not become public.
			\item Identity and location of family members of infected individual should not become public.
			\item The uploaded personal data should not be stored on servers for indefinite time.
		\end{itemize} 

	\subsection{Guidelines for the Non-User Privacy}
		\begin{itemize}
			\item The identity and location of a non-user should not become public.
			\item The contact tracing application shall not collect any data related to a non-user, for example, Bluetooth name, mobile WiFi hot-spot name, etc.
			\item In case a non-user is found to be infected, then the authorities should request for her consent for retrieving location data from other applications or services installed on her device.
			\item The location data should be retrieved and uploaded by a competitive authority or a trusted application only.
			\item The location data should be stored securely on the government servers and must be properly anonymized before sharing or utilizing it for any analysis.
		\end{itemize}
		
	\subsection{Guidelines for the Business Confidentiality}
		\begin{itemize}
			\item Once an area under the locked-down is disinfected or enough time is elapsed, the infected trails should be wiped from the central database. However, it should not impact the scoring advice given to the contact tracing agencies using the above functionality.
			\item Above areas should not become public knowledge to avoid social boycott or cause loss of business.
		\end{itemize} 
		
	\section{Recommendations for Developer (Government)}
		\subsubsection{Data Collection and Processing}
			\begin{itemize}
				\item Participatory data (Self reporting, Symptom data) should not go to the server, but it can reside on the user device for a medical personal for examination. In case a user is found to be \emph{COVID-19} positive, then the medical personnel can upload the anonymized data for further analytics. Further, the application/system should ensure:
				\begin{itemize}
					\item Integrity and confidentiality of the data
					\item Provenance
					\item Proper handling of outlier cases, for example, handling membership inference attack on sparse dataset/paths
					\item Compliance to existing privacy laws and regulations
				\end{itemize}
				\item Risk score needs to be kept private at the same time appropriate advice to be given and testing should be incentivized as per policies.
				\item The application should allow the end user to blacklist certain locations from getting listed in the path trace, for example, home, office etc.
			\end{itemize}
		
		\subsubsection{Privacy Preserving Analytics}
			\begin{itemize}
				\item As the resources are limited on hand-held devices, therefore, server-side score computation using appropriate math primitives that provide privacy assurance should also be considered, for example, private set intersections, bloom filter.
				\item The application should be able to handle scenarios for outliers, that is, it should not leak identity even if the dataset is sparse or contains some celebrity.
			\end{itemize}
		
		\subsubsection{Transparency and Explainability}
			\begin{itemize}
				\item To increase trust, entire functionality should be open-sourced and the Govt should declare that exactly that code is running, so any further holes can be checked. Like what was done in Israel, there needs to be verification and validation of the compiled code vis-a-vis the open source code that has been originally certified by trusted entities.
				\item All data or graph or path anonymization/redaction technique should be properly explained for expert review.
				\item The algorithm for risk score calculation should be clearly described for better transparency.
				\item For a calculated risk score, the application should explain which factors contributed to the score and in what amount.
				\item Application should disclose data processing model, centralized vs decentralized. Additionally, the application should describe the nature of data persisted on devices, shared with peer-devices and shared with servers for all kinds of processing.
				\item Similarly, mode of data sharing should be clearly stated for device-to-device communication and as well as for device-to-server communication, for example, broadcasting, selected broadcasting, unicasting etc.
			\end{itemize}
		\subsubsection{Usability}
			\begin{itemize}
				\item Graphical User Interface of the tracker application should also be able to handle usability issues. For instance:
					\begin{itemize}
						\item It should not be too complex to be of any use by elderly, people with accessibility issues or people without any formal education. This is important for getting an informed consent from the infected user.
						\item Should not ask too many questions to make the end user uncomfortable or fatigued.
						\item Should provide simple and direct instructions for recommendations or while taking consent.
					\end{itemize}
				\item The application should also provide information related to COVID-19, for example, good practices, symptoms and whom to contact in case of emergency.		
			\end{itemize}
		
\section{Concluding Remarks}
	In this work we have brought up privacy concerns of various stakeholders related to contact tracing mobile
	applications. We have briefly analyzed the privacy controls of some prominent applications and their implications in this paper. It is imperative that these applications should provide some tangible incentives to their users while maintaining strict data privacy guarantees. To ensure that we have enunciated a list of privacy guidelines from the perspective of different stakeholders of the application. Also we have listed several recommendations to application developers regarding data collection, processing, transparency, explainability and usability of their application. We hope that these guidelines and recommendations will be useful inputs to the application development teams and lead to the creation of contact tracing applications that see mass adoption.
	
\section{Acknowledgments}
	The authors are grateful to professor Shivaji Sondhi of Princeton University and Riyanka Roy Choudhury, fellow at CodeX, for discussions and support.

\bibliographystyle{ACM-Reference-Format}
\bibliography{privacyguidelines}

\end{document}